\documentclass[]{article}
\usepackage{uai2015stylefiles/proceed2e}

\usepackage{times}
\usepackage{amsmath, amssymb}
\usepackage{graphicx}
\usepackage{algpseudocode}
\usepackage{hyperref}
\usepackage{natbib}
\usepackage{color}
\usepackage{algorithm}
\usepackage{todonotes}

\definecolor{mydarkblue}{rgb}{0,0.08,0.45}
\hypersetup{ %
    pdftitle={},
    pdfauthor={},
    pdfsubject={},
    pdfkeywords={},
    pdfborder=0 0 0,
    pdfpagemode=UseNone,
    colorlinks=true,
    linkcolor=mydarkblue,
    citecolor=mydarkblue,
    filecolor=mydarkblue,
    urlcolor=mydarkblue,
    pdfview=FitH}

\newcommand{\vx}{\mathbf{x}}

\newcommand{\vr}{\mathbf{r}}

\newcommand{\vI}{\mathbf{I}}

\newcommand{\tra}{^{\mathsf{T}}}

\newcommand{\expect}{\mathbb{E}}
\newcommand{\expectargs}[2]{\mathbb{E}_{#1} \left[ {#2} \right]}

\newcommand{\varL}{\mathcal{L}}
\def\iid{i.i.d.\ }

\newcommand{\N}[2]{\mathcal{N}\!\left(#1,#2\right)}
\DeclareMathOperator{\KLop}{KL}
\newcommand{\KL}[2]{\KLop \left(#1 \middle \| #2 \right)}

\newcommand{\data}{\vx}

\newcommand{\params}{\mathbf{\theta}}
\newcommand{\trans}{T}

\newcommand{\stepsize}{\alpha}

\newcommand{\gradparams}{\nabla L(\params_t, t)}

\newcommand{\entropy}{S}

\newcommand{\jointdist}{p(\params , \data)}
\newcommand{\posterior}{p(\params | \data)}
\newcommand{\subjointdist}[2]{p_{#1}(\params_{#2} , \data)}

\newcommand{\reals}{\mathbb{R}}
\newcommand{\bigo}[1]{\mathcal{O}\left(#1\right)}
\newcommand{\trace}[1]{\text{Tr}\left[#1\right]}
\newcommand{\loss}{L(\params)}

\makeatletter
\renewcommand*{\@fnsymbol}[1]{\ensuremath{\ifcase#1\or \dagger\or \ddagger\or
   \mathsection\or \mathparagraph\or \|\or **\or \dagger\dagger
   \or \ddagger\ddagger \else\@ctrerr\fi}}
\makeatother

\title{Early Stopping is Nonparametric Variational Inference}

\author{ {\bf Dougal Maclaurin $^\dagger$} \\
Harvard University
\And
{\bf David Duvenaud \thanks{ \hspace{1em}Equal contributors.}}  \\
Harvard University
\And
{\bf Ryan P. Adams}   \\
Harvard University
}

\begin{document}

\maketitle

\begin{abstract}
We show that unconverged stochastic gradient descent 
can be interpreted as a procedure that samples from a nonparametric variational approximate posterior distribution.
This distribution is implicitly defined as the transformation of an initial distribution by a sequence of optimization updates. 
By tracking the change in entropy over this sequence of transformations during optimization, we form a scalable, unbiased estimate of the variational lower bound on the log marginal likelihood.
We can use this bound to optimize hyperparameters instead of using cross-validation.
This Bayesian interpretation of SGD suggests improved, overfitting-resistant optimization procedures, and gives a theoretical foundation for popular tricks such as early stopping and ensembling.
We investigate the properties of this marginal likelihood estimator on neural network models.
\end{abstract}

\section{Introduction}

In much of machine learning, the central computational challenge is optimization: we try to minimize some training-set loss with respect to a set of model parameters.
If we treat the training loss as a negative log-posterior, this amounts to searching for a maximum \emph{a posteriori} (MAP) solution.
Paradoxically, over-zealous optimization can yield worse test-set results than incomplete optimization due to the phenomenon of \emph{over-training}.
A popular remedy to over-training is to invoke ``early stopping'' in which optimization is halted based on the continually monitored performance of the parameters on a separate validation set.
However, early stopping is both theoretically unsatisfying and incoherent from a research perspective: how can one rationally design better optimization methods if the goal is to achieve something ``powerful but not \emph{too} powerful''?
A related trick is to ensemble the results from multiple optimization runs from different starting positions.
Similarly, this must rely on imperfect optimization, since otherwise all optimization runs would reach the same optimum.

\begin{figure}[t]
\vskip 0.434in
\begin{center}
\includegraphics[width=\columnwidth]{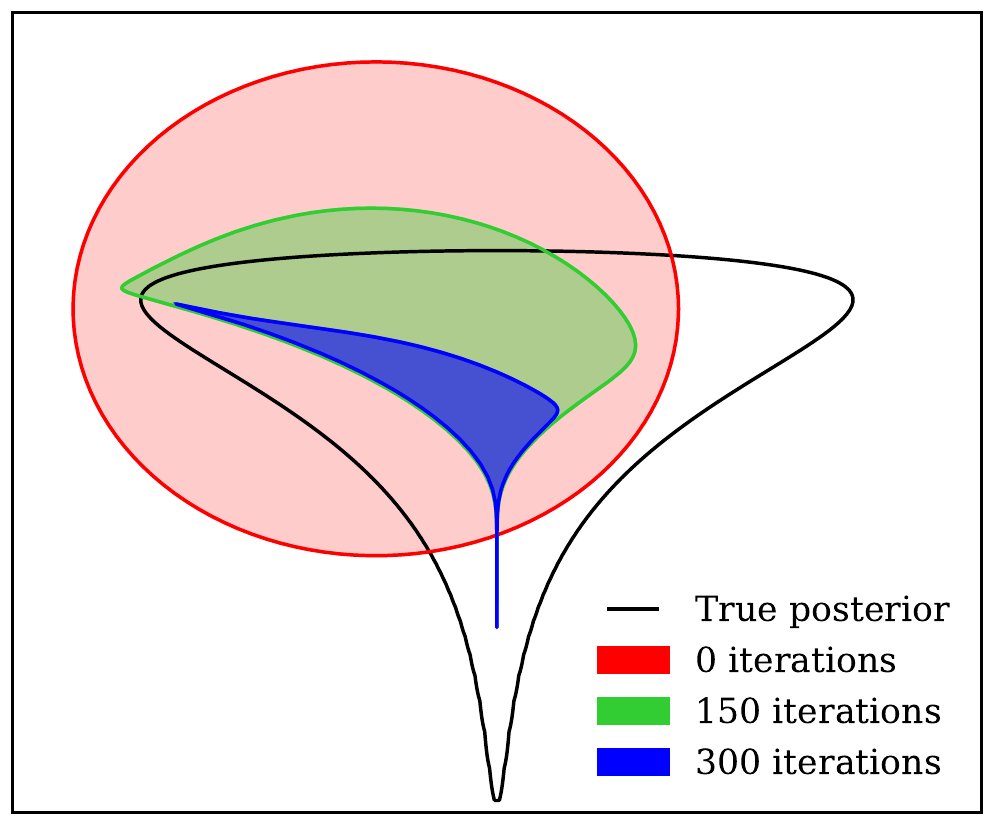}
\vskip -0.043in
\caption{A series of variational distributions implicitly defined by gradient descent on the log-likelihood of the posterior.
Intermediate distributions (green and blue) are implicitly defined by mapping each possible random initial parameters through many iterations of optimization.
  These distributions don't have fixed parametric shape, and will eventually concentrate around the mode.}
\label{fig:cartoon}
\end{center}
\end{figure}

We propose an interpretation of incomplete optimization in terms of variational Bayesian inference, and provide a simple method for estimating the marginal likelihood of the approximate posterior.
Our starting point is a Bayesian posterior distribution for a potentially complicated model, in which there is an empirical loss that can be interpreted as a negative log likelihood and regularizers that have interpretations as priors.
One might proceed with MAP inference, and perform an optimization to find the best parameters.
The main idea of this paper is that such an optimization procedure, initialized according to some distribution that can be chosen freely, generates a sequence of distributions that are implicitly defined by the action of the optimization update rule on the previous distribution.
We can treat these distributions as variational approximations to the true posterior distribution.
A single optimization run for $N$ iterations represents a draw from the $N$th such distribution in the sequence.
Figure \ref{fig:cartoon} shows contours of these approximate distributions on an example posterior.

With this interpretation, the number of optimization iterations can be seen as a variational parameter, one that trades off fitting the data well against maintaining a broad (high entropy) distribution.
Early stopping amounts to optimizing the variational lower bound (or an approximation based on a validation set) with respect to this variational parameter.
Ensembling different random restarts can be viewed as taking independent samples from the variational posterior.

To establish whether this viewpoint is helpful in practice, we ask: can we efficiently estimate the marginal likelihood implied by unconverted optimization?
We tackle this question in section \ref{sec:techintro}.
Specifically, for stochastic gradient descent (SGD), we show how to compute an unbiased estimate of a lower bound on the log marginal likelihood of each iteration's implicit variational distribution.
We also introduce an `entropy-friendly' variant of SGD that maintains better-behaved implicit distributions.

We also ask whether model selection based on these marginal likelihood estimates picks models with good test-time performance.
We give some experimental evidence in both directions in section \ref{sec:experiments}.
A related question is how close the variational distributions implied by various optimization rules approximate the true posterior.
We briefly address this question in section \ref{sec:limitations}.

\subsection{Contributions}
\begin{itemize}
\item We introduce a new interpretation of optimization algorithms as samplers from a variational distribution that adapts to the true posterior, eventually collapsing around its modes.
\item We provide a scalable estimator for the entropy of these implicit variational distributions, allowing us to estimate a lower bound on the marginal likelihood of any model whose posterior is twice-differentiable, even on problems with millions of parameters and data points.
\item In principle, this marginal likelihood estimator can be used for hyperparameter selection and early stopping without the need for a validation set.
We investigate the performance of these estimators empirically on neural network models, and show that they have reasonable properties.
However, further refinements are likely to be necessary before this marginal likelihood estimator is more practical than using a validation set.
\end{itemize}

\section{Incomplete optimization as variational inference}
\label{sec:techintro}
Variational inference \citep{wainwright2008graphical} 
aims to approximate an intractable posterior distribution, $\posterior$, with another more tractable distribution, $q(\params)$.
The usual measure of the quality of the approximation is the Kullback-Leibler (KL) divergence from $q(\params)$ to $\jointdist$.
This measure provides a lower bound on the marginal likelihood of the original model;
applying Bayes' rule to the definition of $\KL{q(\params)}{\posterior}$ gives the familiar inequality:
\begin{align}
\log p(\data)
& \geq - \underbrace{\expectargs{q(\params)}{ -\log \jointdist }}_{\textnormal{\normalsize Energy $E[q]$}}
         \underbrace{- \expectargs{q(\params)}{\log  q(\params)}}_{\textnormal{\normalsize Entropy $S[q]$}} \nonumber \\
& := \varL[q] \label{eq:varbound}
\end{align}
Maximizing $\varL[q]$, the variational lower bound on the marginal likelihood, with respect to $q$ minimizes $\KL{q(\params)}{\posterior}$, the KL divergence from $q$ to the true posterior, giving the closest approximation available within the variational family.
A convenient side effect is that we also get a lower bound on $p(\data)$, which can be used for model selection.

To perform variational inference, we require a family of distributions over which to maximize $\varL[q]$. 
Consider a general procedure to minimize the energy~$(-\log\jointdist)$ with respect to~${\params \in \reals^D}$.
The parameters~$\params$ are initialized according to some distribution~$q_0(\params)$ and updated at each iteration according to a transition operation~${\trans : \reals^D \rightarrow \reals^D}$:
\begin{align}
\params_0 &\sim q_0(\params) \nonumber \\
\params_{t + 1} &= \trans(\params_t), \nonumber
\end{align}
Our variational family consists of the sequence of distributions~$q_0, q_1, q_2, \ldots$,
where~$q_t(\params)$ is the distribution over~$\params_t$ generated by the above procedure.
These distributions don't have a closed form, but we can exactly sample from $q_t$ by simply running the optimizer for $t$ steps starting from a random initialization.

As shown in (\ref{eq:varbound}), $\varL$ consists of an energy term and an entropy term.
The energy term measures how well~$q$ fits the data and the entropy term encourages the probability mass of~$q$ to spread out, preventing overfitting.
As optimization of~$\params$ proceeds from its~$q_0$-distributed starting point, we can examine how~$\varL$ changes.
The negative energy term grows, since the goal of the optimization is to reduce the energy.
The entropy term shrinks because the optimization converges over time.
Optimization thus generates a sequence of distributions that range from underfitting to overfitting, and the variational lower bound captures this tradeoff.

We cannot evaluate $\varL[q_t]$ exactly, but we can obtain an unbiased estimator.
Sampling~$\params_0$ from~$q_0$ and then applying the transition operator~$t$ times produces an exact sample~$\params_0$ from~$q_t$, by definition.
Since $\params_t$ is an exact sample from $q_t(\theta)$, $\log\subjointdist{}{t}$ is an unbiased estimator of the energy term of (\ref{eq:varbound}).
The entropy term is trickier, since we do not have access to the density $q(\params)$ directly.
However, if we know the entropy of the initial distribution, $S[q_0(\params)]$, then we can estimate $S[q_t(\params)]$ by tracking the change in entropy at each iteration, calculated by the change of variables formula. 

To compute how the volume shrinks or expands due to an iteration of the optimizer, we require access to the Jacobian of the optimizer's transition operator, $J(\params)$:
\begin{align}
S[q_{t+1}] - S[q_t] =
  \expect_{q_t(\params_t)} \big[ \log
    \left| J(\params_t) \right| \big] \,.
\end{align}
Note that this analysis assumes that the mapping $T$ is bijective.
Combining these terms, we have an unbiased estimator of $\varL$ at iteration $T$,
based on the sequence of parameters, $\params_0, \ldots, \params_T$, from a single training run:
\begin{align}
\varL[q_T] \approx
  \underbrace{\log \subjointdist{}{T}}_{\textnormal{\normalsize Energy}} +
  \underbrace{\sum_{t=0}^{T-1} \log \left| J(\params_t) \right| + S[q_0]}_{\textnormal{\normalsize Entropy}} \,.
\label{eq:entropy-bound}
\end{align}

\section{The entropy of stochastic gradient descent}
In this section, we give an unbiased estimate for the change in entropy caused by SGD updates.
We'll start with a na\"ive method, then in section \ref{sec:scalable-estimator}, we give an approximation that scales linearly with the number of parameters in the model.

Stochastic gradient descent is a popular and effective optimization procedure with the following update rule:
\begin{align}
\params_{t+1} &=
  \params_t - \stepsize \nabla \loss,
\end{align}
where the $\loss$ the objective loss (or an unbiased estimator of it e.g. using minibatches)
for example ~$-\log\jointdist$, and $\stepsize$ is a `step size' hyperparameter.
Taking the Jacobian of this update rule gives the following unbiased estimator
for the change in entropy at each iteration:
\begin{align}
S[q_{t+1}] - S[q_t] \approx \log \left| I - \stepsize H_t(\params_t)
\right| \label{eq:exact hessian}
\end{align}
where $H_t$ is the Hessian of $-\log\subjointdist{t}{}$ with respect to~$\params$.

Note that the Hessian does not need to be positive definite or even non-singular.
If some directions in $\params$ have negative curvature, as on the crest of a hill, it just means that optimization near there spreads out probability mass, increasing the entropy.
There are, however, restrictions on $\stepsize$.
If ${\stepsize\lambda_i = 1}$, for any $i$, where $\lambda_i$ are the eigenvalues of $H_t$, then the change in entropy will be undefined (infinitely negative).
This corresponds to a Newton-like update where multiple points collapse to the optimum in a single step giving a distribution with zero variance in a particular direction.
However, gradient descent is unstable anyway if ${\stepsize\lambda_{\text{max}} > 2}$, where~$\lambda_{\text{max}}$ is the largest eigenvalue of~$H_t$.
So if we choose a sufficiently conservative step size, such that $\stepsize\lambda_{\text{max}} < 1$,
this situation should not arise.
Algorithm~\ref{alg:sgd-with-estimate} combines these steps into an algorithm that tracks the approximate entropy during optimization.

\begin{algorithm}[t]
   \caption{stochastic gradient descent with entropy estimate}
   \label{alg:sgd-with-estimate}
\begin{algorithmic}[1]
	\State {\bfseries input:}
	Weight initialization scale $\sigma_0$, step size $\stepsize$,
	twice-differentiable negative log-likelihood $L(\params, t)$
	\State {\bfseries initialize} $\params_0 \sim \N{0}{\sigma_0 \vI_D}$
	\State {\bfseries initialize} $\entropy_{0} = \frac{D}{2} (1 + \log 2 \pi) + D \log\sigma_0$
	\For{$t=1$ {\bfseries to} $T$}
		\State $\entropy_{t} = \entropy_{t-1} + \log \left| \vI - \stepsize H_{t-1} \right|$\Comment{Update entropy} \label{step:entropy-update}
		\State $\params_{t} = \params_{t-1} - \stepsize \gradparams$  \Comment{Update parameters}	
   \EndFor
   \State \textbf{output} sample $\params_T$, entropy estimate $\entropy_T$
\end{algorithmic}
\end{algorithm}

So far, we have treated SGD as a deterministic procedure even though, as the name suggests,
the gradient of the loss at each iteration may be replaced by a stochastic
version. Our analysis of the entropy is technically valid if we fix the sequence of stochastic gradients to be the same for each optimization run, so that the only randomness comes from the parameter initialization.
This is a tendentious argument, similar to arguing that a pseudorandom sequence of numbers has only as much entropy as its seed.
However, if we do choose to randomize the gradient estimator differently for each training run
(e.g. choosing different minibatches) then the expression for the change in entropy, Equation \ref{eq:exact hessian}, remains valid as a \emph{lower bound} on the change in entropy and the 
subsequent calculation of $\varL$ remains a true lower bound on the log marginal likelihood.

\subsection{Estimating the Jacobian in high dimensions}
\label{sec:scalable-estimator}
The expression for the change in entropy given by (\ref{eq:exact hessian}) is impractical for large-scale problems since it requires an~$\bigo{D^3}$ determinant computation.
Fortunately, we can make a good approximation using just one or two Hessian-vector products, which can usually be performed in~$\bigo{D}$ time using reverse-mode differentiation \citep{pearlmutter1994fast}.

The idea is that since~$\stepsize\lambda_{\text{max}}$ is small, the Jacobian is actually just a small perturbation to the identity and we can approximate its determinant using traces as follows:
\begin{align}
\log \left| I - \stepsize H \right|
& =    \sum_{i=0}^D \log\left(1 - \stepsize\lambda_i\right) \nonumber\\
& \geq \sum_{i=0}^D \left[- \stepsize\lambda_i 
                        - (\stepsize\lambda_i)^2 \right] \label{eq:logbound} \\
& = - \stepsize \trace{H} - \stepsize^2 \trace{HH}\,.
\end{align}
The bound in (\ref{eq:logbound}) is just a second order Taylor expansion of~$\log(1 - x)$ about~${x = 0}$ and is valid if ${\stepsize\lambda_i < 0.68}$.
As we argue above, the regime in which SGD is stable requires that $\stepsize\lambda_{\text{max}} < 1$, so again choosing a conservative learning rate keeps this bound in the correct direction.
For sufficiently small learning rates, this bound becomes tight.

The trace of the Hessian can be estimated using inner products of random vectors
\citep{bai1996some}:
\begin{align}
\trace{H} = \expectargs{}{\vr^TH\vr}, \qquad \vr \sim \N{0}{I}\,.
\label{eq:approx-log-det}
\end{align}
We use this identity to derive algorithm~\ref{alg:fast-logdet-estimate}.
In high dimensions, the exact evaluation of the determinant in step~\ref{step:entropy-update} should be replaced with the approximation given by algorithm~\ref{alg:fast-logdet-estimate}.

\begin{algorithm}[t]
   \caption{linear-time estimate of log-determinant of Jabobian of one iteration of stochastic gradient descent}
   \label{alg:fast-logdet-estimate}
\begin{algorithmic}[1]
	\State {\bfseries input:}
	step size $\stepsize$, current parameter vector $\params$,
	twice-differentiable negative log-likelihood $L(\params)$
	\State {\bfseries initialize} $\vr_0 \sim \N{0}{\sigma_0 \vI_D}$
	\State $\vr_1 = \vr_0 - \stepsize \vr_0\tra \nabla \nabla L(\params, t)$
	\State $\vr_2 = \vr_1 - \stepsize \vr_1\tra \nabla \nabla L(\params, t)$
	\State $\hat \varL = \vr_0\tra \left( -2 \vr_0 + 3 \vr_1 - \vr_2 \right)$
    \State \textbf{output} $\hat \varL$, an unbiased estimate of a parabolic lower bound on the change in entropy.
\end{algorithmic}
\end{algorithm}

Note that the quantity we are estimating \eqref{eq:exact hessian} is well-conditioned, in contrast to the related problem of computing the log of the determinant of the Hessian itself.
This arises, for example, in making the Laplace approximation to the posterior \citep{mackay1992practical}.
This is a much harder problem since the Hessian can be arbitrarily ill-conditioned, unlike our small Hessian-based perturbation to the identity.

\subsection{Parameter initialization, priors, and objective functions}
\label{sec:priors}
What initial parameter distribution should we use for SGD?
The marginal likelihood estimate given by \eqref{eq:entropy-bound} is valid no matter which initial distribution we choose.
We could conceivably optimize this distribution in an outer loop using the marginal likelihood estimate itself.

However, using the prior distribution has several advantages.
First, it is usually designed to have broader support than the likelihood.
Since SGD usually decreases entropy, starting with a high-entropy distribution
is a good heuristic.

The second advantage has to do with our choice of objective function.
The obvious choice is the (unnormalized, negative) log-posterior, but we can actually use any function we like.
A more sensible choice is the negative log-likelihood.
variational distributions only differ from the
initial distribution to the extent that the posterior differs from the prior.
One nice implication is that the entropy estimate will be exactly correct for parameters that don't affect the likelihood.
Because of these favorable properties, we use these choices for the initial distribution and objective in our experiments.

\section{Designing entropy-friendly optimization methods}
\label{sec:entropy friendly}
SGD optimizes the training loss, not he variational lower bound.
In some sense, if this optimization happens to create a good variational distribution, it's only by accident.
Why not design a new optimization method that produces good variational lower bounds?
In place of SGD, we can use any optimization method for which we can approximate the change in entropy, which in practice means any optimization for which we can compute Jacobian-vector products.

An obvious place to start is with stochastic update rules inspired by Markov Chain Monte Carlo (MCMC). 
Procedures like Hamiltonian Monte Carlo \citep{neal2011mcmc} and Langevin dynamics MCMC \citep{welling2011bayesian} look very much like optimization procedures but actually have the posterior as their stationary distribution.
This is exactly the approach taken by \citet{Bridging14}.
One difficulty with using stochastic updates, however, is that calculating the change in
entropy at each iteration requires access to the current distribution over parameters.
As an example, consider that convolving a delta function with a Gaussian yields an
infinite entropy increase, whereas convolving a broad uniform distribution with a Gaussian
yields only a small increase in entropy. \citet{welling2011bayesian} handle this
by learning a highly parameterized ``inverse model'' which implicitly models the distribution
over parameters. The downside of this approach is that the parameters of this model must be learned in an outer loop.

\begin{figure}
\begin{center}
\includegraphics[width=0.9\columnwidth]{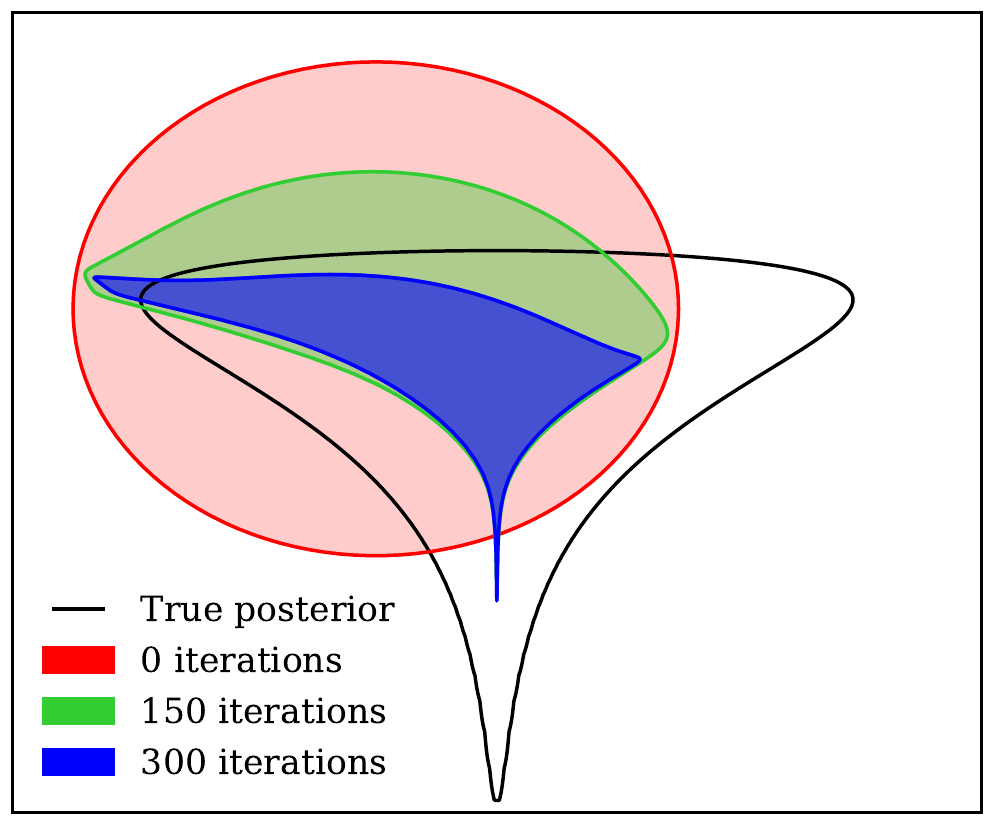}
\caption{The variational distribution implied by the modified, ``entropy-friendly'', SGD algorithm.
Compared to Figure \ref{fig:cartoon}, the variational distributions are slower to collapse into low-entropy filaments, causing the marginal likelihood to remain higher.}
\label{fig:cartoon-fatter}
\end{center}
\end{figure}

Another approach is to try to develop deterministic update rules
that avoid some of the pathologies of update rules like SGD.
This could could be a research agenda in itself, but we give one example here of a modification to
SGD which can improve the variational lower bound.
One problem with SGD in the context of posterior approximation is that
SGD can collapse the variational distribution into low-entropy filaments, shrinking in some directions to be orders of magnitude smaller than the width of the true posterior.
A simple trick to prevent this is to apply a nonlinear, parameter-wise warping
to the gradient, such that directions of very small gradient do not get optimized all the way
to the optimium.
For example, the modified gradient (and resulting modified Jacobian) could be
\begin{align}
g' & = g - g_0 \tanh \left(g / g_0 \right) \\
J' & = \left(1 - \cosh^{-2} (g / g_0) \right) J 
\end{align}
where $g_0$ is a ``gradient threshold'' parameter that sets the scale of this shrinkage.
The effect is that entropy is not removed from parameters which are close to their optimum.
An example showing the effect of this entropy-friendly modification is shown in Figure~\ref{fig:cartoon-fatter}.

\section{Experiments}
\label{sec:experiments}

In this section we show that the marginal likelihood estimate can be used to choose when to stop training, to choose model capacity, and to optimize training hyperparameters without the need for a validation set.
We are not attempting to motivate SGD variational inference as a superior alternative to other procedures;
we simply wish to give a proof of concept that the marginal likelihood estimator has reasonable properties.
Further refinements are likely to be necessary before this marginal likelihood estimator is more practical than simply using a validation set.

\subsection{Choosing when to stop optimization}

\begin{figure}[h!]
\begin{center}
\includegraphics[width=\columnwidth]{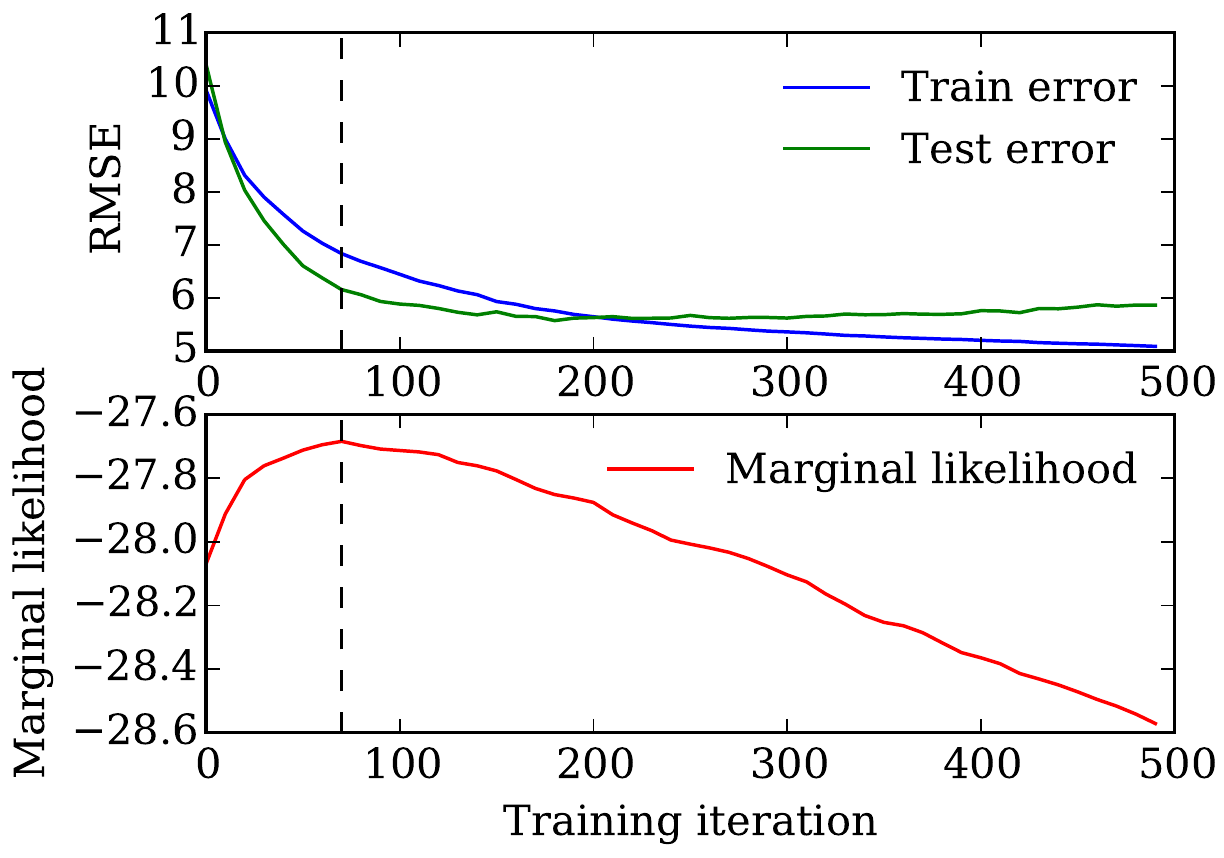}
\vskip -0.1in
\caption{\emph{Top}: Training and test-set error on the Boston housing dataset.
\emph{Bottom}: Stochastic gradient descent marginal likelihood estimates.
The dashed line indicates the iteration with highest marginal likelihood.
The marginal likelihood, estimated online using only the training set, and the
test error peak at a similar number of iterations.}
\label{fig:housing}
\end{center}
\end{figure}

As a simple demonstration of the usefulness of our marginal likelihood estimate, we show that it can be used to estimate the optimal number of training iterations before overfitting begins.
We performed regression on the Boston housing dataset 
using a neural network with one hidden layer having 100 hidden units, sigmoidal activation functions, and no regularization.
Figure \ref{fig:housing} shows overfitting and shows that marginal likelihood peaks at a similar place to the peak of held-out log-likelihood, which is where early stopping would occur when using a large validation set.

\subsection{Choosing the number of hidden units}


The marginal likelihood estimate is also comparable between training runs, allowing us to use it to select model hyperparameters, such as the number of hidden units.

\begin{figure}[h!]
\begin{center}
\includegraphics[width=\columnwidth]{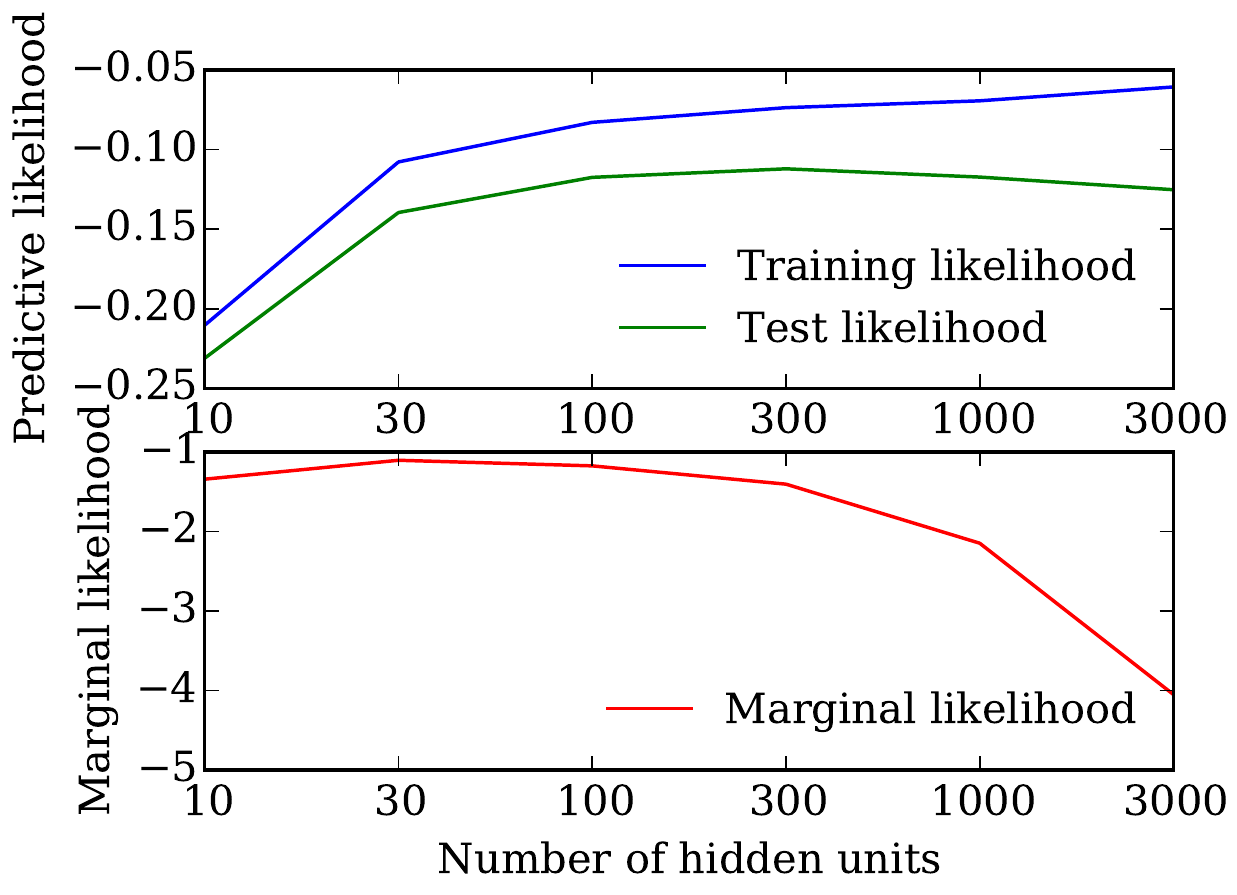}
\vskip -0.1in
\caption{\emph{Top}: Training and test-set likelihood as a function of the number of hidden units in the first layer of a neural network.
\emph{Bottom}: Stochastic gradient descent marginal likelihood estimates.
In this case, the marginal likelihood over-penalizes high numbers of hidden units. 
}
\label{fig:num hiddens}
\end{center}
\end{figure}

Figure \ref{fig:num hiddens} shows marginal likelihood estimates as a function of the number of hidden units in the hidden layer of a neural network trained on 50,000 MNIST handwritten digits.
The largest network trained in this experiment contains 2 million parameters.

The marginal likelihood estimate begins to decrease for more than 30 hidden units, even though the test-set likelihood in maximized at 300 hidden units.
We conjecture that this is due to the marginal likelihood estimate penalizing the loss of entropy in parameters whose contribution to the likelihood was initially large, but were made irrelevant later in the optimization.

\subsection{Optimizing training hyperparameters}

\begin{figure}[h!]
\begin{center}
\includegraphics[width=\columnwidth]{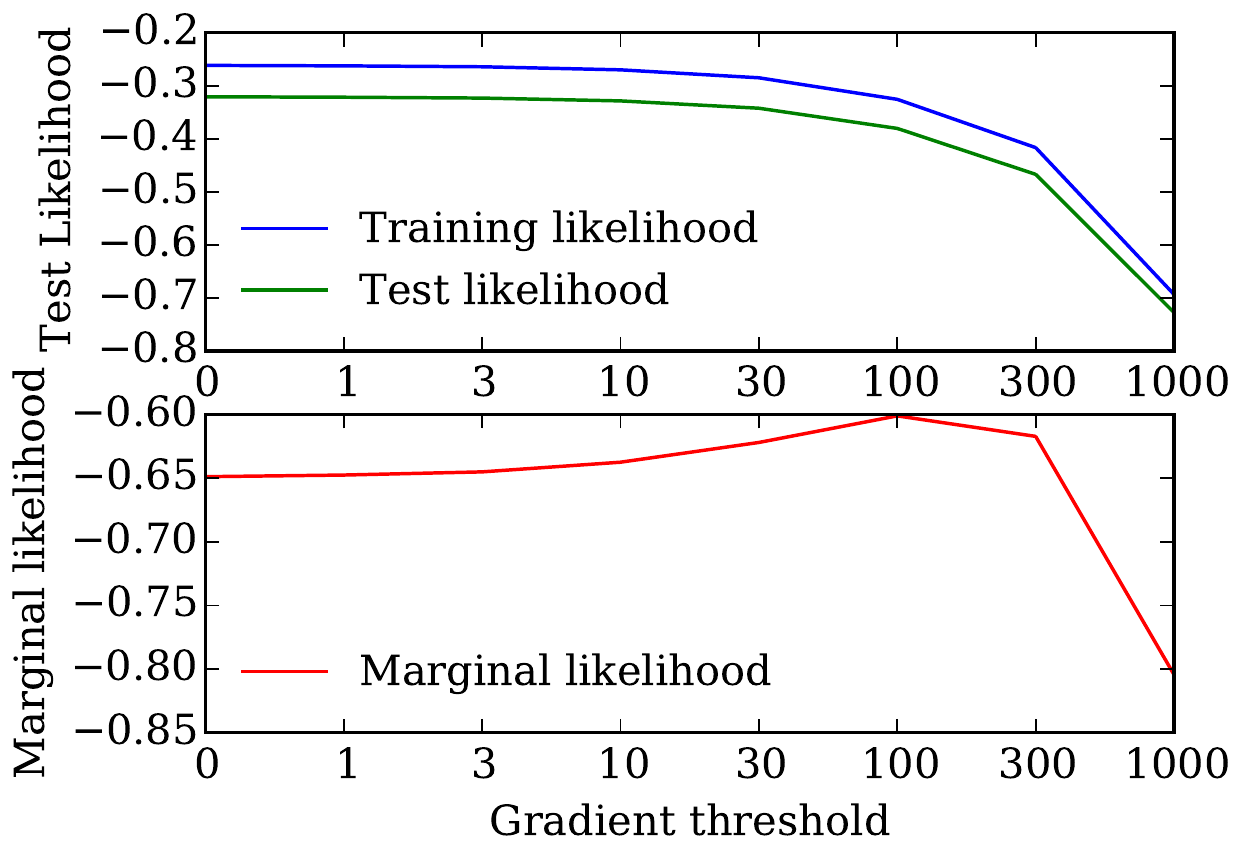}
\vskip -0.1in
\caption{\emph{Top}: Training and test-set likelihood as a function of the gradient threshold.
\emph{Bottom}: Marginal likelihood as a function of the gradient threshold.
A gradient threshold of zero corresponds to standard SGD.
The increased lower bound for non-zero thresholds indicates that the entropy-friendly variant of SGD is producing a better implicit variational distribution.}
\label{fig:threshold}
\end{center}
\end{figure}

We can also use marginal likelihoods to optimize training parameters such as learning rates, initial distributions, or any other optimization parameters.
As an example, Figure \ref{fig:threshold} shows the marginal likelihood estimate as a function of the gradient threshold in the entropy-friendly SGD algorithm from section \ref{sec:entropy friendly} trained on 50,000 MNIST handwritten digits.

As the level of thresholding increases, the training and test error get worse due to under-fitting.
However, for intermediate thresholds, the lower bound increases.
Because it is a lower bound, its increase means that the estimate of the marginal likelihood is becoming \emph{more accurate}, even though the actual model happens to be getting worse at the same time.

\subsection{Implementation details}
To allow easy computation of Hessian-vector products in arbitrary models, we implemented
a reverse-mode automatic differentiation package for Python, available at \url{github.com/HIPS/autograd}.
This package operates on standard Numpy~\citep{oliphant2007python} code, and can differentiate code containing loops, branches, and even its own gradient evaluations.

Code for all experiments in this paper is available at \url{github.com/HIPS/maxwells-daemon}.

\section{Limitations}
\label{sec:limitations}
In practice, the marginal likelihood estimate we present might not be useful for several reasons.
First, using only a single sample to estimate both the expected likelihood as well as the entropy of an entire distribution will necessarily have high variance under some circumstances.
These problems could conceivably be addressed by ensembling, which has an interpretation as taking multiple exact independent samples from the implicit variational posterior.

Second, as parameters converge, their entropy estimate (and true entropy) will continue to decrease indefinitely, making the marginal likelihood arbitrarily small.
However, in practice there is usually a limit to the degree of overfitting possible.
This raises the question: when are marginal likelihoods a good guide to predictive accuracy?
Presumably the marginal likelihood is more likely to be correlated with predictive performance when the
implicit distribution has moderate amounts of entropy.
In section \ref{sec:entropy friendly} we modified SGD to be less prone to produce regions of pathologically low entropy, but a more satisfactory solution is probably possible.

Third, if the model includes a large number of parameters that do not affect the predictive likelihood, but which are still affected by a regularizer, their convergence will penalize the marginal likelihood estimate even though these parameters do not affect test set performance.
This is why in section \ref{sec:priors} we recommend optimizing only the log-likelihood, and incorporating the regularizer directly into the initialization procedure.
More generally however, entropy could be underestimated if a large group of parameters are initially constrained by the data, but are later ``turned off'' by some other parameters in the model.

Finally, how viable is optimization as an inference method?
Standard variational methods find the best approximation in some class, but SGD doesn't even try to produce a good approximate posterior, other than by seeking the modes.
Indeed, Figure \ref{fig:cartoon} shows that the distribution implied by SGD collapses to a small portion of the true posterior early on, and mainly continues to shrink as optimization proceeds.
However, the point of early stopping is not that the intermediate distributions are particularly good approximations, but simply that they are better than the point masses that occur when optimization has converged.


\section{Related work}

\paragraph{Estimators for early stopping}
Stein's unbiased risk estimator (SURE) \citep{stein1981estimation} provides an unbiased estimate of generalization performance under very broad conditions, and can be used to construct a stopping rule.
\citet{raskutti2014early} derived a SURE estimate for SGD in a regression setting.
Interestingly, this estimator depends on the `shrinkage matrix' $\prod_{t=0}^{T} \left( \vI - \stepsize_t H_T \right)$, which is just the Jacobian of the entire SGD procedure along a particular path.
However, this estimator depends on an estimate of the noise variance, and is restricted to the \iid regression setting.
It's also not clear if these stopping rules could also be used to select other training parameters or model hyperparameters.


\paragraph{Reversible learning} 
Optimization is an intrinsically information-destroying process, since a (good) optimization procedure maps any initial starting point to one or a few final optima.
We can quantify this loss of information by asking how many bits must be stored in order to reverse the optimization, as in \citet{MacDuvAda2015hyper}.
We can think of the number of bits needed to exactly reverse the optimization procedure as the average number of bits `learned' during the optimization.

From this perspective, stopping before optimization converges can be seen as a way to limit the number of bits we try to learn about the parameters from the data.
This is a reasonable strategy, since we don't expect to be able to learn more than a finite number of bits from a finite dataset.
This is also an example of reducing the hypothesis space to improve generalization.

\paragraph{MCMC for variational inference}
Our method can be seen as a special case of \citet{Bridging14}, who showed that any set of stochastic dynamics, even those not satisfying detailed balance, can be used to implicitly define a variational distribution.
However, to provide a tight variational bound, one needs to estimate the entropy of the resulting implicit distribution.
\citet{Bridging14} do this by defining an inverse model which estimates backwards transition probabilities, and then optimizes this model in an outer loop.
In contrast, our dynamics are deterministic, and our estimate of the entropy has a simple fixed form.

\paragraph{Bayesian neural networks}
Variational inference has been performed in Bayesian neural-network models~\citep{graves2011practical, deepGPVar14, Miguel2015pbp}.
\citet{kingma2014efficient} show how neural networks having unknown weights can be reformulated as neural networks having known weights but stochastic hidden units, and exploit this connection to preform efficient gradient-based inference in Bayesian neural networks.

\paragraph{Black-box stochastic variational inference}
\citet{alp2014blackbox} introduce a general scheme for variational inference using only the gradients of the log-likelihood of a model.
However, they constrain their variational approximation to be Gaussian, as opposed to our free-form variational distribution.





\section{Future work and extensions}

\paragraph{Optimization with momentum}
One obvious extension would be to design an entropy estimator of 
momentum-based optimizers such as stochastic gradient descent with momentum, or
refinements such as Adam~\citep{Adam14}.
However, it is difficult to track the entropy change during the updates to the momentum variables.


\paragraph{Gradient-based hyperparameter optimization}
Hyperparameters typically come in two forms:
Regularization parameters and training parameters.
Optimizing marginal likelihood rather than training loss lets us set regularization parameters during training without using a validation set.
The marginal likelihood estimate lets us optimize the variational parameters (training hyperparameters) in an outer loop.
However, optimizing more than a few of these is difficult without gradients.
We could gain access to exact gradients of the variational lower bound with respect to all variational parameters by simply using reverse-mode differentiation.
\citet{domke2012generic, MacDuvAda2015hyper} showed that this can be done in a memory-efficient way for momentum-based learning procedures.
Combining these two procedures would allow one to set all hyperparameters using gradient-based methods without the need for a validation set.

\paragraph{Stochastic dynamics}
One possible method to deal with over-zealous reduction in entropy by SGD would be to add noise to the dynamics.
In the case of Gaussian noise, we would recover Langevin dynamics~\citep{neal2011mcmc}.
However, estimating the entropy becomes much more difficult in this case.
\citet{welling2011bayesian} introduced stochastic gradient Langevin dynamics for doing inference with minibatches.
\citet{ma2013estimating} use Langevin dynamics and a floating temperature to estimate partition functions of graphical models.

More generally, we are free to design optimization algorithms that do a better job of producing samples from the true posterior, as long as we can track their entropy.
The gradient-thresholding method proposed in this paper is a simple first example of a refinement to SGD that maintains a tractable entropy estimate while improving the quality of the intermediate distributions.

\section{Conclusion}

Optimization algorithms with random initializations implicitly define a series of distributions which converge to posterior modes.
We showed that these nonparametric distributions can be seen as variational approximations to the true posterior. 
We showed how to produce an unbiased estimate of this variational lower bound by approximately tracking the entropy change at each step of optimization.

This simple and inexpensive calculation turns standard gradient descent into an inference algorithm, and allows the optimization of hyperparameters without a validation set.
Our estimator is compatible with using data minibatches and scales linearly with the number of parameters, making it suitable for large-scale problems.

\subsection{Acknowledgements}
We are grateful to Roger Grosse, Miguel Hern\'andez-Lobato, Matthew Johnson, and Oren Rippel for helpful discussions.
We thank Analog Devices International and Samsung Advanced Institute of Technology for their support.

\bibliography{references.bib}
\bibliographystyle{uai2015stylefiles/icml2015}

\end{document}